\pdfoutput=1

\documentclass[11pt]{article}

\usepackage[final]{acl}

\usepackage{times}
\usepackage{latexsym}

\usepackage{spverbatim}
\usepackage{graphicx}
\usepackage{booktabs}
\usepackage{multirow}
\usepackage{amsmath}
\usepackage{tikz}
\usepackage{standalone}
\usepackage{dialogue}
\usepackage{makecell}
\usepackage{mdframed}

\usepackage{xcolor}
\usepackage{ulem} 

\definecolor{arancione}{RGB}{255,69,0} 
\definecolor{viola}{RGB}{186,85,211} 

\newcommand{\extInconsinstency}[1]{\textbf{\textcolor{arancione}{#1}}}
\newcommand{\intInconsinstency}[1]{\textbf{\textcolor{viola}{#1}}}

\usepackage[T1]{fontenc}

\usepackage[utf8]{inputenc}

\usepackage{microtype}

\usepackage{inconsolata}

\graphicspath{ {./img/} }

\title{Evaluating Task-Oriented Dialogue Consistency through Constraint Satisfaction}

\author{Tiziano Labruna \and Bernardo Magnini \\
Fondazione Bruno Kessler \\ Via Sommarive, 18, Trento, Italy}

\begin{document}
\maketitle
\begin{abstract}
Task-oriented dialogues must maintain consistency both within the dialogue itself, ensuring logical coherence across turns, and with the conversational domain, accurately reflecting external knowledge.
We propose to conceptualize dialogue consistency as a Constraint Satisfaction Problem (CSP), wherein variables represent segments of the dialogue referencing the conversational domain, and constraints among variables reflect dialogue properties, including linguistic, conversational, and domain-based aspects. To demonstrate the feasibility of the approach, we utilize a CSP solver to detect inconsistencies in dialogues re-lexicalized by an LLM.
Our findings indicate that: (i) CSP is effective to detect dialogue inconsistencies; and (ii) consistent dialogue re-lexicalization is challenging for state-of-the-art LLMs, achieving only a 0.15 accuracy rate when compared to a CSP solver. 
Furthermore, through an ablation study, we reveal that constraints derived from domain knowledge pose the greatest difficulty in being respected.
We argue that CSP captures core properties of dialogue consistency that have been poorly considered by approaches based on component pipelines. 
\end{abstract}

\section{Introduction}
\label{introduction}
Task-oriented dialogue (TOD) systems \citep{mctear2020conversational, louvan-magnini-2020-recent, Balaraman-Survey-21} play a crucial role in human-computer interaction, facilitating seamless communication between users and machines to accomplish specific tasks. A peculiar characteristic of TODs is that they need to maintain consistency at two levels: (i) internally within the dialogue itself, ensuring that information in a turn is consistent with information in other turns, and (ii) consistency with the content of the conversational domain of the dialogue system. Internal consistency is responsible for the coherence of the dialogue, making it possible to maintain a meaningful exchange between the participants. External consistency, on the other hand, allows the dialogue to correctly reflect domain knowledge. In this paper, we investigate how dialogue consistency in TOD can be effectively modeled such that possible violations (i.e., inconsistencies) can be automatically detected. 
\begin{figure}
    \centering
    \begin{verbatim}
R1: N=Taberna A=centre F=spanish P=cheap
R2: N=Espana A=centre F=spanish P=moderate
R3: N=Beirut A=centre F=lebanese P=cheap
    \end{verbatim}
    \begin{mdframed}
    \begin{dialogue}
    \item[U1] I am looking for a restaurant serving \extInconsinstency{Spanish} food.
    \item[S1] There are \intInconsinstency{three} restaurants serving \extInconsinstency{Spanish} food, one is \textbf{cheap} and the other is \textbf{moderate} price range. Which price range would you prefer?
    \item[U2] I am looking for a \textbf{cheap} restaurant in any area that serves \extInconsinstency{Spanish} food.
    \item[S2] \textbf{Beirut} is \textbf{cheap} and serves \extInconsinstency{Lebanese} food. Would you like the location information?
    \end{dialogue}
\end{mdframed}
    \caption{An inconsistent task-oriented dialogue with a  Knowledge Base. Red values indicate internal inconsistencies, purple values indicate external inconsistencies.}
    \label{fig:diag_example}
\end{figure}


Figure \ref{fig:diag_example}, shows a fragment of a Knowledge Base (three restaurants in a city) and a short dialogue in which a user expresses preferences for restaurants serving Spanish food, and the system responds   providing information about available options. There are two inconsistencies in this dialogue: first, at turn S1, the system mentions three restaurants serving Spanish food, which is not consistent with the domain knowledge, where there are two such restaurants (domain inconsistency). Second, at turn S2, the system introduces a Lebanese restaurant, while it would have been expected to mention a Spanish restaurant (dialogue inconsistency).
We assume that a well-formed TOD should not manifest any inconsistency of the type reported in our example. However, while relevant work on evaluating TODs has focused on single dialogue components (e.g., dialogue state tracking \cite{henderson-etal-2014-second}), consistency evaluation has received much less attention. The problem is  even more urgent now that end-to-end approaches \cite{bang2023multitask, lai2023external} are by-passing component evaluations.
 Automatic detection of dialogue inconsistencies is crucial  when dialogues are generated by Large Language Models (LLMs), using few-shot or zero-shot approaches. While LLMs have the capacity to generate TODs without being fine-tuned on training data, it is well known that they are prone to hallucinations \cite{ji2022survey}, which may affect dialogue consistency. Furthermore, in dynamic domains where the conversational context evolves over time \cite{labruna2023addressing, labruna2022finetuning}, maintaining dialogue consistency becomes even more challenging. The possible presence of inconsistencies in TODs \cite{qin2021don} raises the problem of detecting them, which is the topic of the paper.
 
The novel intuition of the paper is to consider dialogue consistency as a kind of \textit{Constraint Satisfaction Problem} (\textit{CSP}). We investigate how to  assess the consistency of a TOD under the following working hypothesis: (i) first, dialogue consistency can be modeled with constraints that need to be respected by appropriate linguistic realizations; (ii) such constraints can be well represented to define a CSP, whose allowed solutions can be identified by a CSP solver; (iii) a TOD is consistent if its linguistic realizations belong to the set of solutions allowed by a CSP solver for that dialogue. In the paper, we discuss how dialogue constraints are defined, how they can be extracted and modeled as a CSP, and how to set up an experimental setting where we can empirically prove that a CSP solver can detect inconsistencies in a dialogue.

The contributions of the paper are the following: (i) we model TOD consistency as CSP: to the best of our knowledge, this is a fully original approach; (ii) we set up a reusable experimental setting where TOD consistency can be automatically evaluated against a CSP solver;\footnote{All resources are publicly available at https://github.com/mwozgpt/tod-csp} (iii) we show that current state-of-the-art LLMs still struggle to solve simple dialogue consistency tasks, which opens to further research in dialogue consistency.

\section{Dialogue Consistency as a Constraint Satisfaction Problem}

In this section, we explore the conceptualization of dialogue consistency in the CSP framework. We first describe the fundamental component of a conversational domain (Section \ref{conversational_domain}), then we elucidate the various constraints that contribute to dialogue coherence (Section \ref{tod_consistency}), encompassing linguistic, dialogic, and domain-based considerations. We finally expound upon the formalization of dialogue constraints as CSPs (Section \ref{csp}), delineating the process of modeling dialogue coherence as a constraint satisfaction task.

\subsection{Conversational Domain}
\label{conversational_domain}
A conversational domain for a TOD refers to the specific topic that the dialogue revolves around, encompassing all the knowledge that is pertinent to the conversation. 
In this context, the conversational domain is typically represented by a domain ontology providing a schema of the concepts (e.g., \textsc{Restaurant}, \textsc{Hotel}, \textsc{Movie}), a  set of slots $S$ (e.g., \textsc{Food}, \textsc{Area}, \textsc{Price}) for the concepts, and the set of values that each slot can assume (e.g., \textsc{Expensive}, \textsc{Moderate}, and \textsc{Cheap} for the \textsc{Price} slot). 
Then, a domain $KB$ comprises a collection of instances for the ontology concepts,  each  consisting of [slot,slot-value] pairs, adhering to the domain ontology schema.


\subsection{Dialogue Consistency}
\label{tod_consistency}

A TOD can be conceptualized as a sequence of conversational turns between a user and a system aimed at achieving a specific goal. Within this framework, ensuring the consistency of the dialogue is crucial for effective communication between the user and the system. We consider three types of constraints, which need to be respected for a dialogue to be consistent: linguistic, dialogic and domain-based constraints.

\paragraph{Linguistic Constraints.} They are necessary to respect general linguistic rules of language, including morpho-syntactic rules (e.g., genre and number agreement) and syntax-based rules (e.g., the correct use of a preposition). For instance, if we are given with the following masked utterance:
\begin{verbatim}
    U: I am looking for a restaurant in <MASK>.
\end{verbatim} 

\noindent
the choice of \textit{center} as substitute to the mask token is valid, while \textit{expensive} would not be suitable, because the preposition \textit{in} is rarely used to introduce a price in English.

\paragraph{Dialogic Constraints.} They  maintain the semantic coherence across successive turns of the dialogue, ensuring that each utterance logically aligns  with the preceding context, thereby facilitating a seamless flow of information. As an example, suppose the following masked dialogue turns:

\begin{verbatim}
U: I would like an Italian restaurant.
S: There is no <MASK> restaurant in the 
   center.
\end{verbatim}

\noindent
Here both \textit{Italian} and \textit{cheap} would be eligible choices from a linguistic point of view, but only \textit{Italian} would maintain the coherence with the previous turn in the dialogue.

\paragraph{Domain Constraints.} They  ensure alignment between the dialogue content and the domain knowledge, thereby maintaining the dialogue's alignment with relevant factual information. Consider, for instance, a $KB$ with the following restaurants:
\begin{verbatim}
R1: N=Mario A=east F=italian P=expensive
R1: N=Napoli A=centre F=italian P=moderate    
\end{verbatim}
\noindent  
And the following piece of masked dialogue:
\begin{verbatim}
U: I am looking for an Italian restaurant 
   in the centre.
S: We have <MASK> restaurants available for 
   your preferences.
\end{verbatim} 
\noindent
Then, the only admissible choice for the masked token would be \textit{one}, as selecting any other number would introduce an inconsistency with the information provided in the $KB$.

\subsection{Dialogue Consistency as CSP}
\label{csp}

A CSP \cite{kumar1992algorithms} imposes certain conditions on a finite set of variables through constraints. Each variable has a finite set of possible values, known as its domain, and constraints define which combinations of values are allowed for specific subsets of the variables. A constraint can be given either explicitly, by enumerating the tuples allowed, or implicitly, e.g., by an algebraic expression. The solution of a CSP is an instantiation of all the variables for which all the constraints are satisfied. A CSP is solvable if it has at least one solution, otherwise it is unsolvable or overconstrained.

The hypothesis of this paper is that the dialogue constraints outlined in Section \ref{tod_consistency} can be modeled as CSPs. 
Intuitively, variables are the portions of the dialogue that need to be constrained (i.e., the <MASK> tokens in our examples), while the range of possible values for the variables are expressed, either explicitly or implicitly, in a domain $KB$ for that dialogue. The CSP task consists of selecting variable assignments that adhere to linguistic, dialogic, and domain constraints.
To formalize this notion, consider a dialogue $d_i$ for which $n$ variables (i.e., masked tokens) $x_1, x_2, \ldots, x_n$ have been defined. Let $D_i$ denote the domain of possible values for variable $x_i$; let $\mathcal{C}$ be the set of constraints (i.e.,  linguistic, dialogic, and domain constraints) over the dialogue $d_i$, and let  $c$ represent a single constraint in $\mathcal{C}$. The CSP task is to determine if there exists an assignment $A = \{(x_1, a_1), (x_2, a_2), \ldots, (x_n, a_n)\}$ with $a_i \in D_i$ for $1 \leq i \leq n$, such that $A$ satisfies all constraints in $\mathcal{C}$. This problem can be formulated as follows:
\[
\text{\textit{Satisfies}}(\{(x_1, a_1), (x_2, a_2), \ldots, (x_n, a_n)\}, C_j) 
\]
\[
\quad \forall C_j \in \mathcal{C}
\]
where $\text{\textit{Satisfies}}(A, C_j)$ denotes the binary relationship between an assignment $A$ and a constraint $C_j$, indicating whether the assignment satisfies the constraint.

\begin{figure*}
  \centering \includegraphics[height=0.29\textwidth]{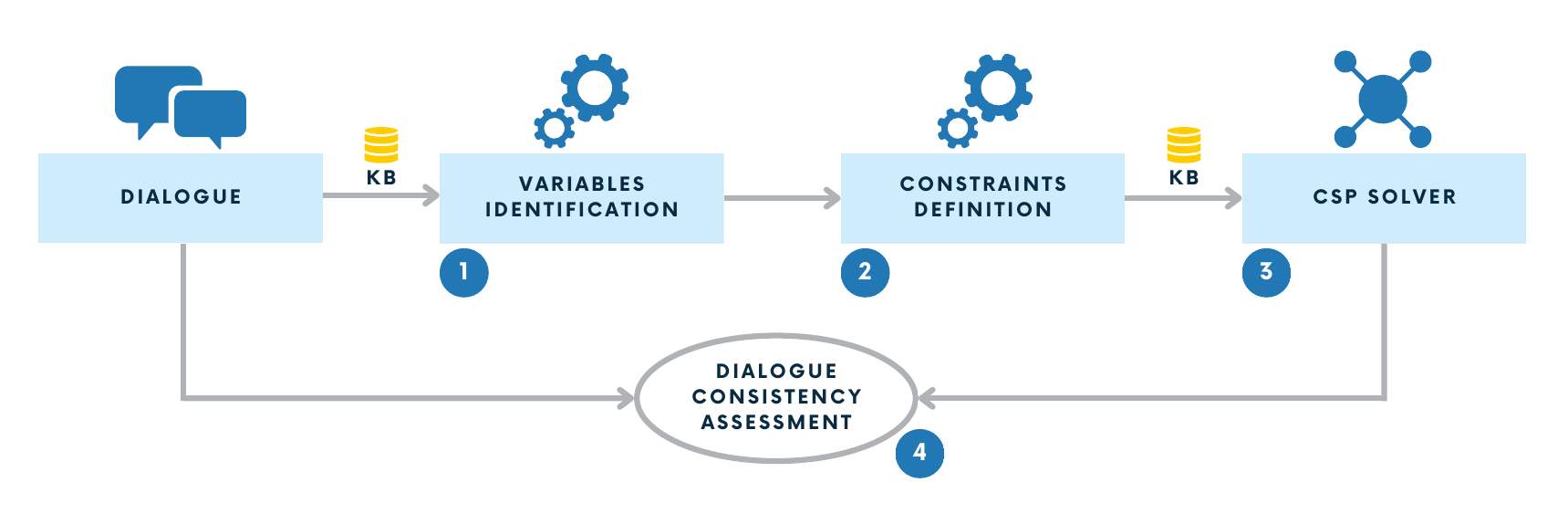}
  \caption{Overview of the CSP-based  methodology applied to TOD consistency.}
  \label{fig:process}
\end{figure*}

\section{Methodology}
\label{methodology}
This section outlines the process of modeling a TOD as a CSP, and then to assess the dialogue consistency using a CSP solver. The assessment involves three key steps for a   $[d, kb]$ pair, where $d$ is a dialogue  and $kb$ is a Knowledge Base: (1) identification of variables within the dialogue $d$ (Section \ref{id_var}); (2) extraction of dialogue constraints and construction of a CSP solver for the  $[d, kb]$ pair (Section \ref{constraints}); and (3) application of the CSP solver to determine if the dialogue $d$ represents a feasible solution with respect to the defined constraints (Section \ref{assess}). These phases of the methodology are illustrated in Figure \ref{fig:process}.


\subsection{Identifying Dialogue Variables}
\label{id_var}

At step 1 (see Figure \ref{fig:process}), we consider a TOD $d$ and a $kb$ (i.e., a set of entities described by slot-value pairs) related to the conversational domain of the dialogue. We do not assume any particular dependency between $d$ and $kb$: $d$ could be either fully covered by $kb$ (i.e., all mentions of slot values in $d$ are present in $kb$), only partially covered, or not covered at all. We consider text portions in $d$ referring to the conversational domain as potential CSP variables: a text portion referring to a slot value or mentioning amounts of instances in $kb$. The rationale is that both slot values and instance amounts are elements that better characterize a TOD and are responsible for its consistency. In our example in Figure \ref{fig:diag_example}, we will obtain the following variables with their assignments:

$[x_1 = Spanish]$, $[x_2 = three]$, $[x_3 = Spanish]$, $[x_4 = cheap]$ ... $[x_{10} = Lebanese]$.



\subsection{Extracting Dialogue Constraints}
\label{constraints}

We have now established a set $\mathcal{X}$ of variables $x_1, x_2, ..., x_n$, where each variable $x_i$ can assume a value either from the slot values or from instance amounts described in $kb$. Moving to   step 2  in Figure \ref{fig:process}, we now extract the set of constraints $\mathcal{C}$ over the values that can be assigned to $\mathcal{X}$ variables. 
We consider the three categories of constraints introduced in Section \ref{tod_consistency}: linguistic, dialogic, and domain-based constraints. 



\paragraph{Extracting linguistic constraints.}
 We model linguistic constraints as the need for a variable derived from a slot value to match the semantic type of its slot type.
 For instance, given the utterance \textit{I am looking for a restaurant at \textsc{$x_1$}}, the  value of the variable \textsc{$x_1$} must belong to the \textsc{Area} type. More precisely, $C1$ is defined as follows:
\[C1: x_1 \in V\]
where $V$ is the set of values belonging to the same slot type as the original value.
Constraint $C1$, is meant to avoid that a variable can assume values that are semantically non valid. For instance, avoiding that \textsc{$x_1$=north} can be assigned to a \textsc{Food}, as in \textit{I am looking for a restaurant at \textsc{indian}}, which is ungrammatical in English.

\paragraph{Extracting dialogic constraints.}
There are two dialogic constraints that we currently consider. $C2$ for ensuring that variables mentioning the same slot value in $d$ are assigned to the same value. $C3$ for ensuring that variables with the same semantic type occurring in the same utterance are assigned to different values. 
Given the turn \textit{U: I want an $x_1$ restaurant. S: There are 3 restaurant that serve $x_2$}, we define $C2$ as follows:
\[C_2: x_1 = x_2\]

\noindent
where the aim is to keep internal coherence across the dialogue turns. 
Given the utterance \textit{We have $x_1$, $x_2$, or $x_3$ restaurants.}, we define $C3$ as:
\[
C_3: x_1 \neq x_2, \quad x_1 \neq x_3, \quad x_2 \neq x_3
\]

\noindent
which captures non redundancy at the utterance level.

\paragraph{Extracting domain-based constraints.} There are three domain-based constraints that we currently consider. All of them are meant to guarantee consistency between the number of instances mentioned in $d$ and the actual number of instances present in $kb$. We distinguish three cases: $C4$ covers the cases when an utterance in $d$ states that there are no instances in $kb$; $C5$ covers the cases where it is stated that there is at least one instance; and $C6$ the cases where there are exactly $n$ instances.

As for $C4$, consider an utterance indicating no results for a search: \textit{There are no restaurants serving $x_1$ food}, assuming that there are no restaurants with \textsc{[Food=$x_1$]} in $kb$. For this utterance, $C4$ is defined as:
\[C4: \neg \exists i \in KB \text{ with values } x_1\]
implying that the variable $x_1$ can not assume a value that is present in an instance of the $KB$.

As for $C5$, consider the utterance:
 \textit{We have many $x_1$ restaurants at $x_2$}, where at least one restaurant with \textsc{[Food=$x_1$]} and \textsc{[Area=$x_2$]} is supposed to exist in $kb$. For this utterance, $C5$ is defined as:
\[C5: \exists i \in KB \text{ with values } x_1, x_2\]
imposing the existence of at least one instance with values $x_1$ and $x_2$.

Finally, for $C6$, consider the utterance \textit{There are $x_1$ restaurants at $x_2$}. We define the constraint as:
\[C6: |\{i \in KB \text{ with value }x_2\}| = x_1\]
to check that the number of instances with value $x_2$ is exactly equal to $x_1$. 

\subsection{Assessing Dialogue Consistency}
\label{assess}

Once all variables and constraints for a dialogue $d$ are identified, a CSP solver computes possible solutions for the variables in $d$ given $kb$ (step 3 in Figure \ref{fig:process}). If one of these solutions matches the variable assignments in $d$, we consider $d$ consistent with respect to $kb$ (step 4 in Figure \ref{fig:process}).
For example, in the dialogue and $kb$ illustrated in Figure \ref{fig:diag_example}, the variable assignments do not match any CSP admissible solution. Specifically, variable assignment $[x_2 = three]$ violates $C6$, referring to an incorrect number of Spanish instances in $kb$, and variable $[x_{10} = Lebanese]$ violates $C2$, as it does not maintain coherence with the previous turns.
If the CSP solver finds at least one solution, the variable assignments in the dialogue must match one of those solutions, ensuring all constraints are followed. On the other hand, if no solution is found with respect to $kb$, the variable assignments should be empty or contain values not in $kb$ to ensure consistency. These aspects will be further explored in the experiments discussed in Section \ref{experiments}.

\section{Experimental Setting}
\label{experiments}
In this section, we present the experimental setup used to assess dialogue consistency through a CSP solver. We describe the general setting and the purposes of the experiments (Section \ref{exp:setting}), the dataset utilized (Section \ref{mwoz}), the KBs associated to each dialogue (Section \ref{extract_kb}), the tools employed for constraint satisfaction (Section \ref{minizinc}), the language model used for dialogue generation (Section \ref{gpt}), the baselines against which we compare our results (Section \ref{baselines}) and finally, the evaluation metrics that have been used (Section \ref{metrics}). 


\begin{table}
\centering
\begin{tabular}{l c c}
\toprule
\textbf{Dataset} & \textbf{\# dialogues} & \textbf{\# variables}\\
\midrule
\textsc{All} & 131 & 768\\
\midrule
\textsc{0 sol.} & 56 & 403\\
\textsc{1 sol.} & 16 & 58\\
\textsc{2-10 sol.} & 27 & 143\\
\textsc{11-100 sol.} & 18 & 95\\
\textsc{101+ sol.} & 14 & 69\\
\bottomrule
\end{tabular}
\caption{Dialogue distribution based on the number of solutions provided by the CSP solver.}
\label{tab:diag_sol}
\end{table}

\subsection{Purposes and General Setting}
\label{exp:setting}

The purpose of the experiments is to check the feasibility of the CSP-based approach described in Section \ref{methodology} for detecting dialogue inconsistencies. Our focus is not on optimizing the performance of the CSP solver but rather on investigating critical aspects of the process in a realistic setting. Several steps are involved in this process:

\begin{enumerate}
    \item Initially, we require dialogue-knowledge base (\(d\)-\(kb\)) pairs. As for  dialogues \(d\), we utilize MultiWoz \cite{han2020multiwoz} dialogues, which are already annotated for dialogue state tracking, enabling precise identification of variables within the dialogue. From an annotated MultiWoz dialogue \(d\), we derive a de-lexicalized version \(d_{delex}\), where dialogue content is replaced with CSP variables.
    \item Additionally, for each dialogue, we derive a knowledge base (\(kb\)) from the MultiWoz ontology, allowing variation in both the size and type of instances.
    \item With \(d_{delex}\) and \(kb\) established, the next step involves generating variable assignments that can be assessed via a CSP solver. To produce dialogues with potential realistic inconsistencies, we employ a large language model (LLM). The LLM is tasked with re-lexicalizing the variables (i.e., substituting slot-values to CSP variables) in \(d_{delex}\), considering the provided \(kb\). The LLM prompt is illustrated in Appendix \ref{sec:appendix-prompt}. This re-lexicalization process aims to maximize correctness while adhering to all implicit dialogue constraints.
    \item Finally, the re-lexicalized dialogue \(d_{relex}\) produced by the LLM serves as a variable assignment and is compared with the solutions of the CSP solver on the same \(d\)-\(kb\) pair to produce a consistency score.
\end{enumerate}




\subsection{MultiWOZ Dataset}
\label{mwoz}

The experimental data was sourced from the MultiWOZ 2.3 dataset \cite{han2020multiwoz}, a widely used benchmark for TOD systems comprising more than ten thousand conversations between a user and a system, covering various domains such as restaurants, hotels, or attractions. For our experiments, we focus on restaurant-related dialogues from the MultiWOZ dataset. In total we consider 131 dialogues with 768 total de-lexicalizations (i.e., CSP variables), as shown in the first row of Table \ref{tab:diag_sol}. In addition, Table \ref{tab:diag_sol} categorizes the dataset into groups based on the number of solutions identified by MiniZinc (see Section \ref{minizinc}) for each dialogue.

\begin{table}
\centering
\begin{tabular}{l c c}
\toprule
\textbf{Constraint} & \textbf{\# variables} & \textbf{\% coverage} \\
\toprule
\textsc{C1} & \textbf{768} & \textbf{1.00} \\
\textsc{C2} & 686 & 0.89 \\
\textsc{C3} & 108 & 0.14 \\
\textsc{C4} & 9 & 0.01 \\
\textsc{C5} & 394 & 0.51 \\
\textsc{C6} & 197 & 0.25 \\
\bottomrule
\end{tabular}
\caption{Number of dialogue variables affected by constraints and their proportion.}
\label{tab:constraints}
\end{table}

\subsection{Knowledge Base}
\label{extract_kb}
The $kb$ employed in the experiments are sourced from the MultiWOZ database. Specifically, for each dialogue $d$ in MultiWOZ, we selected a pertinent instance from the global MultiWOZ KB that aligns with the content of the dialogue. This ensures both relevance and coherence between the dialogue and the associated domain information. Additionally, to introduce variability in the composition of the dialogue $kb$, we randomly sampled a set of $n$ instances from the global MultiWOZ KB, where $n$ is a randomly generated number between 0 and 8. This approach ensures a diverse range of instances in the dialogue $kb$ while constraining the total number of instances to a maximum of 9, facilitating efficient prompting of the $kb$ to the LLM.

\subsection{MiniZinc Constraint Solver}
\label{minizinc}

As for CSP solver, we use MiniZinc \cite{nethercote2007minizinc}, an open-source constraint programming language specifically designed for modeling and solving constraint satisfaction problems. We employed MiniZinc to obtain solutions satisfying the dialogue constraints  for ourevaluation purposes.
MiniZinc provides a high-level modeling language that allows users to express problem constraints and objectives. It supports a wide range of constraint types, which make it suitable for modeling diverse problem domains.
 Among MiniZinc's suite of solvers, we leveraged Chuffed \cite{chu2018chuffed}, a state-of-the-art solver known for its efficiency in solving CSPs through time optimization, especially advantageous for addressing complex and large-scale optimization problems.

\subsection{GPT-3.5-Turbo Language Model}
\label{gpt}

For dialogue re-lexicalization, we employed the GPT-3.5-Turbo language model, a member of the OpenAI GPT family \cite{achiam2023gpt}, specifically designed to perform well in conversational contexts. GPT-3.5-Turbo was prompted with both ($d_{delex}$) and its associated $kb$. This comprehensive input served to guide the model to produce dialogues that adhere to the implicit constraints, thereby ensuring dialogue coherence and adherence to the domain. We utilized GPT-3.5 for inference in zero-shot mode (see Appendix \ref{sec:appendix-prompt}), without any fine-tuning, leveraging the API version dated "2023-05-15" with a temperature setting of 0.9 to ensure balanced exploration and exploitation during dialogue generation.

\subsection{Baselines}
\label{baselines}
We introduce two dialogue re-lexicalization baselines, for a comparative analysis with GPT. The first baseline (\textsc{Random}), produces a dialogue $d_{relex}$ where variables in $d_{delex}$ are randomly assigned to slot values present in the $kb$. The second baseline (\textsc{Most Frequent}) produces a dialogue $d_{relex}$ where variables in $d_{delex}$ are assigned to the most frequent value observed in the $kb$. By contrasting our evaluation results with these baselines, we gain insights into the efficacy of our approach in capturing and assessing dialogue consistency.

\begin{table}
\centering
\begin{tabular}{l c c}
\toprule
\textbf{Dataset} & \textbf{GCA} & \textbf{VCA}\\
\midrule
\textsc{Random} & 0.01 & 0.06\\
\textsc{Most Frequent} & 0.01 & 0.10\\
\textsc{GPT} & \textbf{0.15} & \textbf{0.27}\\
\bottomrule
\end{tabular}
\caption{Global and variable consistency for dialogues re-lexicalized by GPT compared to the \textsc{Random} and \textsc{Most Frequent} baselines.}
\label{tab:baselines}
\end{table}

\subsection{Evaluation Metrics}
\label{metrics}
Global Consistency Accuracy (GCA) and Variable Consistency Accuracy (VCA) are the two metrics used to evaluate the adherence of a dialogue to a specific set of constraints. Given a re-lexicalized dialogue $d_{relex}$ where variables are assigned to values, GCA measures the overall accuracy of the assignments for each variable. The average GCA is calculated as the proportion of dialogues that fully comply with all defined constraints:
\[ GCA = \frac{\sum_{i=1}^{N} \left( \prod_{j=1}^{M} \textit{Satisfies}(A_i, C_j) \right)}{N} \]
where \(N\) is the total number of dialogues, and \(\textit{Satisfies}(A_i,C_j)\) is a binary indicator function that returns 1 if and only if all variable assignments in dialogue \(d_i\) comply with the constraint $j$, 0 otherwise.
On the other hand, VCA assesses the assignment accuracy on individual variables within the dialogue. We compare the dialogue assignment to the solutions of the CSP solver and find the most similar solution; then, we count how many variable assignments coincide with the assignments of the most similar solution. We formally define VCA as follows:
\[ VCA = \frac{\sum_{i=1}^{N} \lvert \textit{CorrectAssignments}(d_i) \rvert}{M} \]

\noindent
where \(N\) is the total number of dialogues, \(M\) is the total number of variables in the dialogues, and \(\textit{CorrectAssignments}(d_i)\) are the variable assignments in dialogue \(d_i\) that coincide with the assignments of the most similar solution provided by the CSP solver.
GCA and VCA provide insights into the ability of the dialogue generation system to maintain coherence and fidelity to the underlying domain knowledge while generating responses. Higher values of GCA and VCA indicate better performance in terms of dialogue quality and consistency.


\section{Results}


Table \ref{tab:constraints} presents the impact of each constraint on the variables in the dataset, detailing the percentage of variables influenced by each constraint. This shows that $C1$ (i.e., assigned values need to respect the semantic type of the variable) applies to all variables in the dataset, while $C4$ (no instances in $kb$) applies only nine time in total. 
Table \ref{tab:baselines} compares the global and variable consistency in dialogues re-lexicalized by GPT with the \textsc{Random} and \textsc{Most Frequent} baselines. GPT dialogues exhibit significantly higher global and variable consistency compared to the baseline datasets.
Table \ref{tab:diag_consistency} assesses GCA and VCA for GPT dialogues across various CSP solution groups. Results show that dialogues with more solutions tend to have higher GCA and VCA scores, while the model is not able to recognize and address the 0 solution cases.

Table \ref{tab:ablation} presents the results of an ablation study, where we systematically remove each constraint one by one and analyse their impact on GCA and VCA for each configurations. Results show that the most critical constraint is $C6$ (i.e., exact match with number of $kb$ instances).
Additionally, we conducted experiments where groups of constraints were collectively removed to observe their influence on the dialogue generation process, confirming that domain-based constraints are more critical.

\begin{table}
\centering
\begin{tabular}{l c c}
\toprule
\textbf{Dataset} & \textbf{GCA} & \textbf{VCA}\\
\midrule
\textsc{0 sol.} & 0.0 & 0.0\\
\textsc{1 sol.} & 0.31 & 0.48\\
\textsc{2-10 sol.} & 0.22 & 0.53\\
\textsc{11-100 sol.} & 0.22 & 0.55\\
\textsc{101+ sol.} & \textbf{0.36} & \textbf{0.70}\\
\bottomrule
\end{tabular}
\caption{Assessment of global and variable consistency for re-lexicalized dialogues across solution groups.}
\label{tab:diag_consistency}
\end{table}

\section{Discussion}
The experiment results shed light on several key aspects of consistency assessment for TODs. First, comparing GPT and the two baselines (\textsc{Random} and \textsc{Most Frequent}) on re-lexicalized dialogues, we note the better quality achieved by the GPT model (see Table \ref{tab:baselines}), both in term of GCA and VCA. GPT can effectively re-lexicalize dialogues that more closely adhere to the defined constraints.
Furthermore, the assessment of global and variable consistency across different solution groups reveals interesting patterns (see Table \ref{tab:diag_consistency}). Dialogues with a higher number of solutions tend to exhibit higher levels of consistency, indicating that the model performs better when presented with more options to fulfill constraints. At the other extreme, the model is not able to address cases where no feasible solution exists, as it always provides an attempt of assignment for the variables. This finding emphasizes the importance of considering the richness and diversity of CSP solutions, as they have a strong impact on the quality and consistency of re-lexicalized dialogues.
Additionally, analysing the distribution of constraints on the dialogue variables, reveals significant variations (see Table \ref{tab:constraints}), with certain constraints exerting a stronger influence than others. The ablation study provides valuable insights into the impact of the different constraints on dialogue re-lexicalization. Excluding domain constraints, in particular, leads to significantly higher GCA and VCA scores, indicating the critical role of domain-specific knowledge in shaping dialogue coherence and relevance (see Table \ref{tab:ablation}). This suggests that recent LLMs may not effectively leverage the provided $kb$, highlighting an area for potential improvement in future iterations of language model training and dialogue re-lexicalization techniques. 
Our experiments have shown that modeling and assessing dialogue consistency through CSP is both feasible and challenging. We were able to highlights both strengths and weaknesses of dialogue generation and to discern which constraints are met and which are not, gaining insight into the specific features and challenges inherent in this process. 

\begin{table}
\centering
\begin{tabular}{l c c}
\toprule
\textbf{Constraint} & \textbf{GCA} & \textbf{VCA} \\
\midrule
\makecell[l]{\textsc{all except C1}} & 0.15 & 0.31 \\
\makecell[l]{\textsc{all except C2}} & 0.15 & 0.27 \\
\makecell[l]{\textsc{all except C3}} & 0.15 & 0.29 \\
\makecell[l]{\textsc{all except C4}} & 0.16 & 0.30 \\
\makecell[l]{\textsc{all except C5}} & 0.15 & 0.32 \\
\makecell[l]{\textsc{all except C6}} & \textbf{0.21} & \textbf{0.48} \\
\midrule
\makecell[l]{\textsc{all except} \\ \textsc{dialogic}} & 0.15 & 0.30 \\
\makecell[l]{\textsc{all except} \\ \textsc{domain}} & \textbf{0.23} & \textbf{0.56} \\
\bottomrule
\end{tabular}
\caption{Ablation study: global and variable consistency under different constraint configurations.}
\label{tab:ablation}
\end{table}

\section{Related Work}

TOD systems have been extensively investigated in NLP. \cite{allen2001architecture}. 
Recent research has explored the use of neural network architectures for dialogue state tracking \cite{wu2020gcdst, zhao2021effective} and policy learning \cite{su2016line, liu2017iterative}.
%
Several metrics have been proposed to assess the performance of TOD systems, including task completion rates, user satisfaction scores, and objective measures for system components, such as precision, recall, and F1-score \cite{chen2017survey, santhanam2019towards, deriu2021survey}. Recent studies have emphasized the importance of holistic evaluation frameworks that consider multiple aspects of dialogue quality \cite{zhang2021d}.

Maintaining consistency and coherence in dialogues is essential for effective communication between users and dialogue systems. Previous research has investigated various approaches to ensure dialogue coherence, including coherence modeling \cite{cervone2018coherence}, and coherence-based response generation \cite{cervone2020dialogue}, aiming to enhance the naturalness and fluency of generated dialogues.
Finally, several studies have explored the application of CSPs to language. These include early attempts to ensure coherence in generated text \cite{kibble2004optimizing}, model preposition lexicalization using constraints \cite{moriceau2004constraint}, guide lexical choices through constraints \cite{mckeown1997floating}, and treat context-sensitive utterance generation as a CSP \cite{popescu2009constraint}.


\section{Conclusion}
In this paper, we have introduced a novel approach to assess dialogue consistency in the context of TODs using a metric based on Constraint Satisfaction. In our approach, variables represent de-lexicalized segments of the dialogue and constraints reflect linguistic, conversational, and domain-based properties of TODs. Our experiments have demonstrated the feasibility of this approach, enabling us to effectively identify and quantify inconsistencies present in the dialogues.
An interesting side-effect of our investigation is the observation that state-of-the-art LLMs often introduce numerous inconsistencies when tasked with re-lexicalizing dialogues. These inconsistencies primarily concern domain knowledge adherence, resulting in an overall accuracy of only 0.15 at the dialogue level. 
Our study highlights the potential of CSP-based methodologies in evaluating dialogue consistency and identifying areas for improvement in automated dialogue generation systems. Future research should further explore the application of CSP in this domain and investigate strategies to enhance the coherence of LLM-generated dialogues, particularly in applications with strong domain knowledge requirements.

\section{Limitations}

Our study is subject to several limitations that warrant consideration. Firstly, the process of defining constraints for dialogue consistency assessment is complex and multifaceted. While we have delineated several constraints in this study, the TOD landscape is vast, and additional constraints may need to be identified and incorporated to capture a broader range of dialogue scenarios accurately. Each constraint is formulated based on our current understanding of the phenomena, acknowledging that further investigations may uncover additional constraints. Additionally, we also consider implementation feasibility, as certain constraints may require more extensive implementation efforts to detect. Moreover, the selection and prioritization of constraints inherently involve subjective judgment, and achieving consensus on the most relevant constraints for a given dialogue domain may pose a challenge.

Secondly, while we employed a state-of-the-art Large Language Model (LLM) for dialogue generation and consistency assessment, the performance of alternative language models remains unexplored. Investigating the effectiveness of various LLM architectures, pre-training strategies, or fine-tuning approaches could provide valuable insights into their suitability for TOD tasks.

Furthermore, while our methodology endeavors to be as generalizable as possible, it is important to acknowledge that nuances in dialogue structures and domain-specific knowledge may exist across different datasets, and there may still be aspects of dialogue consistency that our approach may not fully capture. Exploring additional datasets spanning diverse domains and languages could offer a more comprehensive understanding of dialogue consistency challenges and the efficacy of our proposed methodology.

\bibliography{main}

\appendix

\section{Appendix A: GPT prompt}
\label{sec:appendix-prompt}

Below is an instruction that outlines a task, along with a Knowledge Base containing domain-specific information to be utilized, and a dialogue for you to work on. Return a response that effectively fulfills the task.\\\\
\#\#\# Instruction:\\
Fill in the [MASK] placeholders in the dialogue based on the information provided in the Knowledge Base. Provide the updated dialogue exactly as it was given, but with the placeholders replaced by the appropriate values for each turn in the dialogue. If a turn does not contain any placeholders, leave the sentence unchanged. Turns should start with either User or System. Be aware of leaving blank spaces before punctuation as in the original (e.g. Hi , instead of Hi,)\\\\
\#\#\# Knowledge Base: \\
Restaurant \#1 - Area: centre, Food: british, Price: moderate\\
Restaurant \#2 - Area: west, Food: european, Price: expensive\\\\
\#\#\# Dialogue: 
\begin{dialogue}
  \speak{User} I 'm looking for a restaurant serving \textbf{[MASK]} food in any area .
  \speak{System} There are no \textbf{[MASK]} restaurants in the area .
  \speak{User} Well can I get the phone number to a \textbf{[MASK]} restaurant ?
  \speak{System} Restaurant one seven serves \textbf{[MASK]} food . Their phone number is 01223 355166 . Can I help you with anything else ?
  \speak{User} That 's it , goodbye
  \speak{System} your welcome good bye
\end{dialogue}
\#\#\# Response: 
\begin{dialogue}
  \speak{User} I 'm looking for a restaurant serving \textbf{european} food in any area .
  \speak{System} There are no \textbf{european} restaurants in the area .
  \speak{User} Well can I get the phone number to a \textbf{british} restaurant ?
  \speak{System} Restaurant one seven serves \textbf{british} food . Their phone number is 01223 355166 . Can I help you with anything else ?
  \speak{User} That 's it , goodbye
  \speak{System} your welcome good bye
\end{dialogue}

\end{document}